                                                          \documentclass[a4paper, 10pt, conference]{ieeeconf}      
\usepackage{FG2019}

\FGfinalcopy 

\IEEEoverridecommandlockouts                              
\usepackage{cite}
\usepackage{graphicx}
\usepackage{hyperref}
\usepackage{float} 
\usepackage{subfigure} 
\usepackage{url}
\usepackage{multicol}
\usepackage{multirow}
\def\FGPaperID{125} 
\title{\LARGE \bf
\emph{LRW-1000}: A Naturally-Distributed Large-Scale Benchmark \\ for Lip Reading in the Wild
}

\author{\parbox{16cm}{\centering
{\large Shuang Yang\footnote{$^*$ Authors make equal contributions to this work.}$^{*1}$, Yuanhang Zhang$^{*2}$, Dalu Feng$^{*1,2}$, Mingmin Yang$^{*4}$, Chenhao Wang$^{2}$, Jingyun Xiao$^{2}$, Keyu Long$^{2}$, Shiguang Shan$^{1,2,3}$, Xilin Chen$^{1,2}$}\\
{\normalsize
$^1$ Key Laboratory of Intelligent Information Processing of Chinese Academy of Sciences (CAS), Institute of Computing Technology, CAS, Beijing 100190, China\\
$^2$ University of Chinese Academy of Sciences, Beijing 100049, China\\
$^3$ CAS Center for Excellence in Brain Science and Intelligence Technology\\
$^4$ Huazhong University of Science and Technology}}
\thanks{$^*$ Authors make equal contributions to this work. 
}}

\begin{document}
\IEEEoverridecommandlockouts\pubid{\makebox[\columnwidth]{978-1-7281-0089-0/19/\$31.00~\copyright{}2019 IEEE \hfill}
\hspace{\columnsep}\makebox[\columnwidth]{ }}

\ifFGfinal
\thispagestyle{empty}
\pagestyle{empty}
\else
\author{Anonymous FG 2019 submission\\ Paper ID \FGPaperID \\}
\pagestyle{plain}
\fi
\maketitle
\bibliographystyle{plain}

\begin{abstract}

   Large-scale datasets have successively proven their fundamental importance in several research fields, especially for early progress in some emerging topics. In this paper, we focus on the problem of visual speech recognition, also known as lip-reading, which has received increasing interest in recent years. We present a naturally-distributed large-scale benchmark for lip-reading in the wild, named \emph{LRW-1000}, which contains $1,000$ classes with $718,018$ samples from more than $2,000$ individual speakers. Each class corresponds to the syllables of a Mandarin word composed of one or several Chinese characters. To the best of our knowledge, it is currently the largest word-level lip-reading dataset and also the only public large-scale Mandarin lip-reading dataset. This dataset aims at covering a ``natural'' variability over different speech modes and imaging conditions to incorporate challenges encountered in practical applications. It has shown a large variation in this benchmark in several aspects, including the number of samples in each class, video resolution, lighting conditions, and speakers' attributes such as pose, age, gender, and make-up. Besides providing a detailed description of the dataset and its collection pipeline, we evaluate several typical popular lip-reading methods and perform a thorough analysis of the results from several aspects. The results demonstrate the consistency and challenges of our dataset, which may open up some new promising directions for future work.

\end{abstract}

\section{INTRODUCTION}
  Visual speech recognition, also known as lip-reading, is a task of recognizing the speech content in a video only based on visual information. It has been demonstrated that incorporating visual information in audio-based speech recognition systems can bring obvious performance improvements, especially in cases where multiple speakers are present or the acoustic signal is noisy \cite{audio_visual2, audio_visual3}. 
  
  The common procedure of lip-reading involves two steps: analyzing motion information in the given image sequence and transforming this information into words or sentences. This procedure links lip-reading to two closely related fields: audio-based speech recognition and action recognition, both of which relies on a similar analyzation to an input sequence to obtain predicted results. 
  However, currently there exists a large performance gap between lip-reading and these two closely related tasks. One main reason is that there were few large-scale lip-reading datasets in the past, which was likely a major obstacle to the progress in lip-reading. 
  
  Fortunately, with the development of deep learning technologies, some researchers have begun to collect large-scale data for lip-reading in recent years using deep learning tools. Existing public datasets can be divided into two categories: word-level dataset and sentence-level dataset. We focus on word-level lip-reading in this paper. One outstanding benchmark is the \emph{LRW}\cite{VGG-LRW} dataset proposed in 2016, which has $500$ classes and displays substantial diversity in speech conditions. In addition, all the videos in this dataset are of fixed size and length, which provides much convenience to the community. The best performance on this dataset in terms of Top-$1$ classification accuracy has reached as high as $83\%$ in merely two years.
  Some other popular word-level lip-reading datasets include \emph{OuluVS}\cite{ouluvs1} and \emph{OuluVS2}\cite{ouluvs2}, which were released in 2009 and 2015 respectively. There are $10$ classes in both datasets and the state-of-the-art model has achieved an accuracy of more than $90\%$ on both datasets. These exciting and encouraging results mark a significant and praiseworthy improvement in lip-reading. However, lip-reading in natural or ``in-the-wild" settings remains challenging due to the large variations in the practical real-world environment. Meanwhile, these appealing results also call for more challenging datasets to trigger new progresses and inspire novel ideas for lip-reading.
  
  \begin{figure*}[h]
    \centering
    \includegraphics[scale=0.8]{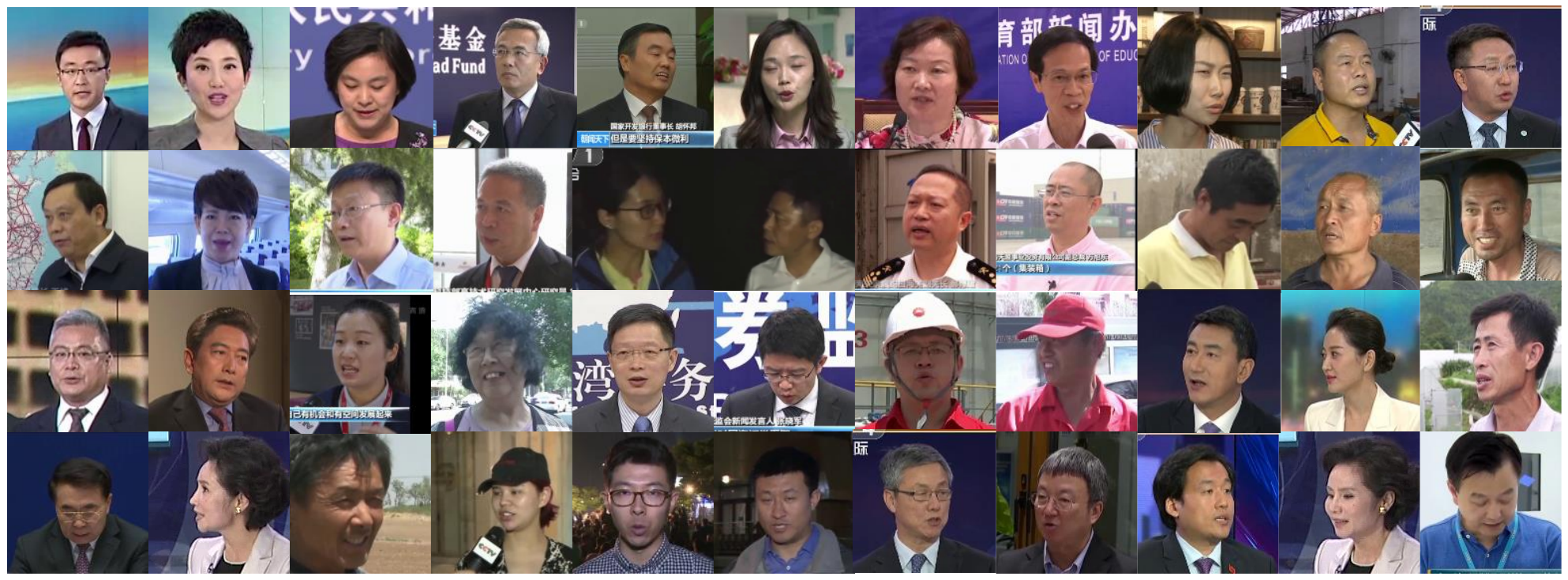}
    \caption{Examples of speakers in our dataset, which show a large variation of speech conditions, including lighting conditions, resolution, speaker's age, pose, gender, and make-up etc.}
    \label{fig_face_examples}
    \vspace{-0.4cm}
  \end{figure*}
  To this end, we collect a naturally-distributed large-scale dataset for lip-reading in the wild. The contributions in this paper are summarized as follows.
  
  Firstly, we present a challenging $1,000$-class lip-reading dataset. Each class corresponds to the syllables of a Mandarin word which is composed of one or several Chinese characters. The labels are also provided in the format of English letters and so anyone who knows English could understand and use the data. In total, there are $718,018$ samples from more than $2000$ speakers, with over $1$ million Chinese character instances covering $286$ Chinese syllables. To the best of our knowledge, this database is currently the largest word-level lip-reading dataset and also the only one large-scale Mandarin lip-reading dataset.
  
  Secondly, our benchmark aims to provide naturally-distributed data to the community, highlighted by the following properties: (a) it contains large variations in speech conditions, including lighting conditions, resolution of videos, and speaker's attribute variations in pose, speech rate, age, gender, make-up and so on, as shown in Fig.~\ref{fig_face_examples}; (b) some classes are allowed to contain more samples than some others, which is consistent with the actual case that some words indeed occur more frequently than others; (c) samples of the same word are not limited to a previously specified length range to allow different speech rates. These three properties make this dataset very consistent with practical settings. 
  
  Thirdly, we provide a comprehensive comparison of the current popular lip-reading methods and perform a detailed analysis of their performance in several different settings to analyze the effect of different factors on lip-reading, including the performance with respect to image scales, word's length, speaker's pose and the model capacity on naturally-distributed data. The results demonstrate the consistency and the challenges of our benchmark, which may lead to some new inspirations to the related research communities.  
\section{Related Work}
\begin{table*}[h]
    \setlength{\abovecaptionskip}{-0.5cm}
    \setlength{\belowcaptionskip}{-0.cm}
    \caption{A Summary of Existing Well-known Word-level Lip-reading Datasets}
    \vspace{1em}
    \label{dataset_table}
    \begin{center}
        \begin{tabular}{|c||c|c|c|c|c|c|c|c|}
            \hline
            \textbf{Datasets} &  \textbf{\# of Classes} & \textbf{\# of Speakers} & \textbf{Resolution} & \textbf{Pose} & \textbf{Envir.} & \textbf{Color/Gray} & \textbf{Year} & \textbf{Best Acc.}\\
            \hline
            \hline
            AVICAR \cite{AVICAR} & 10 & 100 & - & Controlled & In-car & Gray & 2004 & 37.9\% \cite{avicar_best}\\
            \hline
            AVLetters \cite{AVLetters} & 26 & 10 & Fixed $80 \times 60$ & Controlled & Lab & Gray & 2002 &  65.13\% \cite{AV_6513}\\
            \hline
            OuluVS \cite{ouluvs1} & 10 & 20 & Fixed $80 \times  60$ & Controlled & Lab & Color & 2009 &  91.4\% \cite{VGG-LRW}\\
            \hline 
            OuluVS2 \cite{ouluvs2} & 10 & 53 & Fixed (6 different sizes) & Controlled & Lab & Color & 2015  & 95.6\% \cite{SP_bmvc}\\ %
            \hline
            LRW \cite{VGG-LRW} & 500 & $>1000$ & Fixed $256 \times 256$ & Natural & TV & Color & 2016  & 83.0\% \cite{COMBINGING}\\
            \hline
            \textbf{\emph{LRW-1000}} & 1000 & $>2000$ & Naturally distributed & Natural & TV & Color & 2018 &  38.19\%\\
            \hline
        \end{tabular}
    \end{center}
    \vspace{-0.5cm}
\end{table*}

In this section, we provide an overview of current word-level lip-reading datasets, followed by a survey of state-of-the-art methods targeting at lip-reading.

\subsection{Word-level Lip-reading Datasets}  
Some well-known word-level lip-reading datasets are summarized in Table~\ref{dataset_table}. All these datasets have contributed greatly to the progress of lip-reading. In this part, we will give a brief review of these well-known datasets shown in the table.

\emph{AVICAR} \cite{AVICAR} and \emph{AVLetters} \cite{AVLetters} were proposed in $2004$ and $2002$ respectively and were widely used at an early period. The words in these two datasets are composed by $10$ digits and $26$ letters from $100$ speakers and $10$ speakers respectively. These two datasets has provided an important and initial impetus for early progress in automatic lip-reading.

\emph{OuluVS} \cite{ouluvs1}, released in $2009$, consists of $10$ phrases spoken by $20$ subjects with $817$ sequences in total. This dataset provides cropped mouth region sequences, which brings much convenience to related researchers. However, the average number of samples in each class is merely $81.7$, which is not enough to cover the various conditions in practical applications.

\emph{OuluVS2} \cite{ouluvs2}, released in $2015$, extends the number of subjects in \emph{OuluVS} to $53$. The speakers are recorded from five fixed different views: frontal, profile, $30^{\circ}$, $45^{\circ}$ and $60^{\circ}$. One major difference compared with \emph{AVLetters} and \emph{OuluVS} is that \emph{OuluVS2} contains several different viewpoints, which makes it more difficult than the above three datasets and is therefore widely used in previous lip-reading studies. However, the viewpoints are all fixed in this dataset, and also, there are few variations beyond the view conditions. 

\emph{LRW} \cite{VGG-LRW}, an appealing large-scale lip-reading dataset released in 2016, contains $500$ classes with more than a thousand speakers. The videos are no longer posed videos recorded in controlled lab environments as above, but are extracted from TV shows and thus cover a large variation of speech conditions. This remains a challenging dataset until now and has been widely used by most existing lip-reading methods. However, one pre-defining setting of this dataset is that all the words are ensured to have a roughly equal duration and each class is specified to contain roughly the same number of samples. This setting leads to a gap between the data and practical applications because word frequencies and speech rates are actually not uniform in the real world. We believe that if a model learned from data which has a natural diversity over these two points can still achieve good performance, it should also perform well when applied to practical applications.

Although there have been many English lip-reading datasets as listed above, there are very few Mandarin lip-reading datasets available up to now. With the rapid development of scientific technologies, automatic lip-reading of any language would definitely catch more and more researchers' attention over time. Therefore, we hope \emph{LRW-1000} could fill a part of the gap for automatic lip-reading of Mandarin.

\subsection{Lip reading Methods}
Automated lip-reading has been studied in the computer vision fields for decades. Most early methods focus on designing appropriate hand-engineering features to obtain good representations. Some well-known features include the Discrete Cosine Transform (DCT), active appearance model (AAM),  motion history image (MHI), Local Binary Pattern (LBP)  and optical flow , to name a few. 
With the rapid development of deep learning technologies, more and more work began to perform end-to-end recognition with the help of deep neural networks (DNN). 
According to the types of the front-end network, modern lip-reading methods can be roughly divided into the following three categories. 

(1) \emph{Fully 2D CNN based: }   
Two-dimensional convolution has been proved successful in extracting representative features in image-based recognition tasks. With this inspiration, some early lip-reading work \cite{2dCNN_lr}, \cite{cnn_hmm}, \cite{lstm5} try to obtain a discriminative representation of each frame individually with some pretrained 2D CNN models, such as theVGGNet \cite{vgg} and residual networks \cite{resnet}. 
One representative work is the multi-tower structure proposed by Chung and Zisserman in \cite{lstm5}, where each tower takes a single frame or a $T$-channel image as input with each channel corresponding to a single frame in grayscale. The activations from all the towers are concatenated to produce the final representation of the whole sequence. This multi-tower structure has been proved effective by appealing results on the current challenging dataset \emph{LRW}.   

(2) \emph{Fully 3D CNN based: } 
One direct reason for the wide use of 3D convolutional layers in lip-reading has much to do with the success of 3D CNN in action recognition \cite{C3D}. One popular method whose front-end network is completely based on 3D convolution is the LipNet model \cite{LIPNET}. It contains three 3D convolutional layers which transform the raw input video into spatial-temporal features and feed them to the following gated recurrent units (GRUs) to generate the final transcription. The effectiveness has been proved by its remarkable performance on the public dataset which has surpassed professional human lip-readers by a large margin. 

(3) \emph{Mixture of 2D and 3D convolution: }
The regular 2D spatial convolutional layers have been proved to be effective in extracting discriminative features in the spatial domain, while spatio-temporal convolutional layers are believed to be able to better capture the temporal dynamics in a sequence. For this reason, some researchers have begun to combine the advantages of the two types to generate even stronger features. In \cite{COMBINGING}, Stafylakis and Tzimiropoulos proposed to combine a spatio-temporal convolutional layer with a 2D residual network to produce the final representation of the sequence. It has achieved the state-of-the-art performance on \emph{LRW} with an accuracy of $83\%$.

In this paper, we evaluate each of the above state-of-the-art approaches on our proposed benchmark and present a detailed analysis of the results which may provide some inspirations for future research.

\section{Data Construction}
\begin{figure}[t]
  \centering
  \includegraphics[scale=0.48]{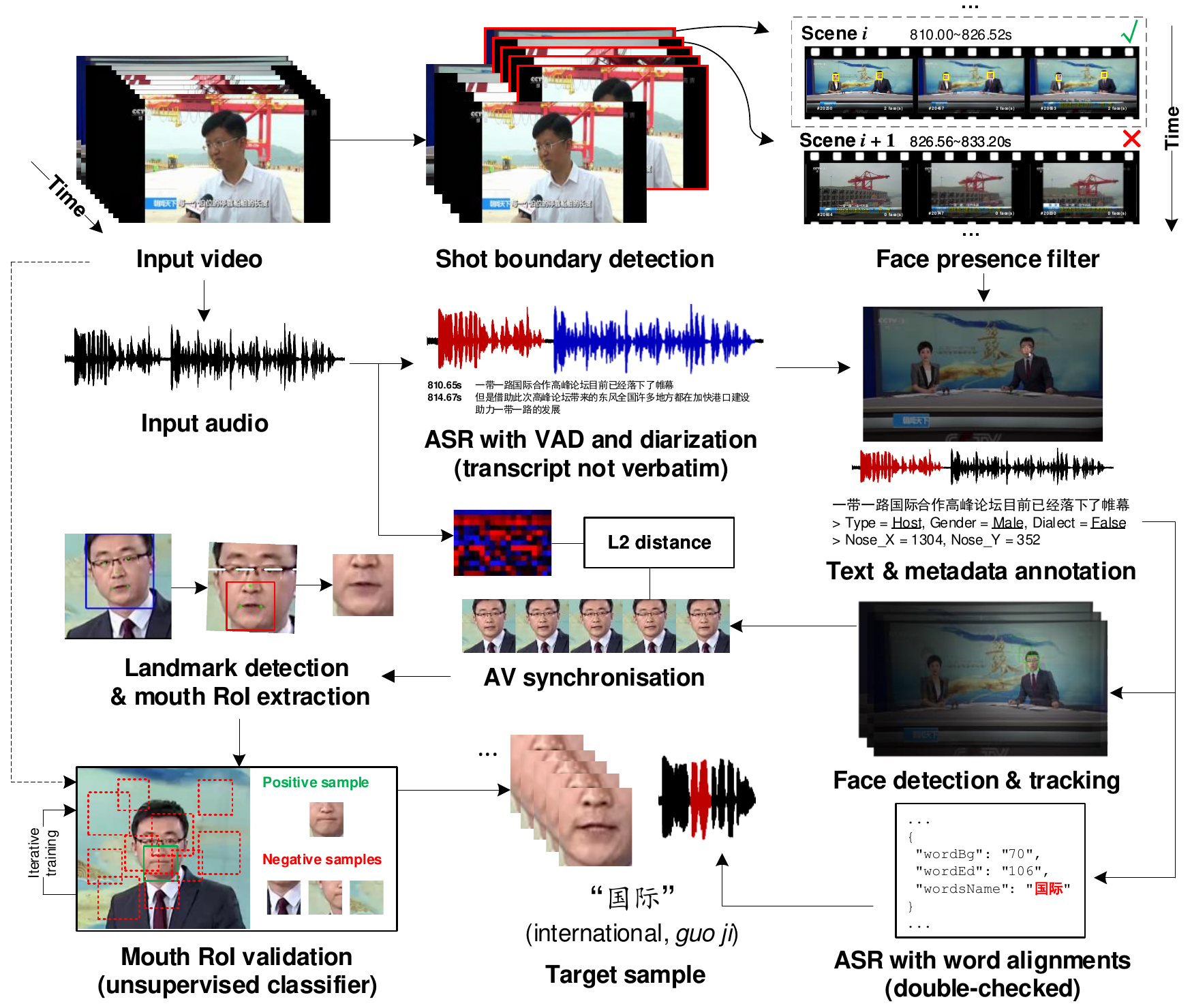}
  \setlength{\abovecaptionskip}{-0.5cm} 
  \setlength{\belowcaptionskip}{-0.5cm} 
  \caption{Pipeline to generate samples in our dataset.}
  \vspace{-0.5cm}
  \label{fig_pipeline}
\end{figure}

In this section, we describe the pipeline for collecting and processing the \emph{LRW-1000} benchmark, as shown in Fig. \ref{fig_pipeline}. We first present the choice of television programs from which the dataset was created and then provide details of the data preprocessing procedures, which interleave automatic process with manual annotation and extra filtering efforts to make the data consistent for research.

\subsection{Program Selection and Data Collection}
In our benchmark, all collected programs are either broadcast news or conversational programs with a focus on news and current events. To encourage the diversity of speakers and speech content, we select programs from both regional and national TV stations, covering a wide range of male and female TV presenters, guests, reporters and interviewees who speak Mandarin or dialectal Chinese. The final program list is composed by $26$ broadcast sources with $51$ programs and yields more than $500$ hours of raw videos over the two-month data collection period. This large range endows the data with a nearly full coverage of commonly used words and a natural diversity in several aspects as in practical applications.

The broadcast collection described above is retrieved daily through an IPTV streaming service in China, hosted by Northeastern University. It produces $25$ fps recordings in H.264 encoding, with $1.5$ to $7.5$Mbps video bitrates and $128$ to $160$Kbps audio bitrates. The video resolution is $1920\times 1080$ for high-definition channels and $1024\times 576$ for standard-definition channels. This makes our data cover a wide range of scales. Since the source videos were recorded through cable TV and re-encoded in real-time, they may contain temporal discontinuities which appear as frozen frames or artifacts. We clip each video up to the first occurrence of such abnormality and feed the obtained segment to subsequent procedures.

\subsection{Shot Boundary Detection}
We firstly employ a shot boundary detector by comparing the color histograms of adjacent frames. Within each detected shot, we choose three evenly spaced frames and perform face detection with a multi-view face detector in the SeetaFaceEngine2 toolkit\cite{seetaface}. If none of them contains a face larger than $20\times 20$ pixels, we dismiss the shot as not containing any potential speakers. What is worth noting is that although we deliberately set a low minimum size of the candidate faces to closely mimic the in-the-wild setting, statistics still show that there are very few samples with lip resolution below $20\times 20$, as shown in Fig.~\ref{fig_scale_distribution}.

\subsection{Annotations, Face Detection, and Face Tracking}
Most Chinese TV programs have no verbatim subtitles, so we create rough transcripts of the videos with the commercial iFLYREC speech recognition service, time-aligned at the sentence level. This process automatically detects voiced segments in the audio track and diarizes it by different speakers. We then isolate sentences which are within shots retained in the previous stage, and manually annotate each video clip with the active speaker's position, gender, exact endpoints of the speech, and also the speech content.
Finally, to further refine the manually-checked text annotations, a more robust ASR tool by iFLYTEK is used to produce very faithful transcripts of the utterance which are compared again with the manually checked transcripts. After several rounds of interleaved mannual and automatic check, the final annotation is believed to be accurate enough for the final use.

To associate each utterance with the corresponding speaker's face, we use the landmark detector in SeetaFaceEngine2 on the first frame and check by comparing the coordinates of each detected face with the manual annotation. Then, a kernelized correlation filter (KCF) tracker is utilized to the selected face in the given duration to obtain the whole speaking sequence. During the tracking process, we perform automatic validation of the tracking quality every $15$ frames with the CNN-based face detector in SeetaFaceEngine2.

\subsection{Audio-to-Video Synchronization}
After the above process, we check for the synchronization issues and find that similar to \cite{lrs3} \cite{vgg_lrs}, the audio and video streams in a few collected videos may be out of sync, with the largest offset being less than one second. To tackle this problem, 
we introduce the SyncNet model in \cite{syncnet}, which extracts visual features from $5$ frames of cropped faces using a 3D VGG-M network 
and computes their distance to the MFCC-based audio features. The model searches for offsets within $\pm 15$ frames, attempting to minimize the distance between the two features so that the two modalities are synchronized. We run the model over all the extracted utterances from each video and average the distances across these samples. If the determined offset is greater than $\pm 7$ frames in any clip or the samples do not reach a consent, we will perform shifting of the video stream manually to obtain the final synchronization.

\subsection{Facial Landmark Detection and Mouth Region Extraction}
At this stage, we have obtained face tracks of individuals speaking, as well as synchronized audio with corresponding transcripts. The next step is to extract the mouth regions. We first detect facial landmarks with the SeetaFaceEngine2 toolkit. Using these landmarks, the detected faces are first rotated so that the eyes are barely on a horizontal line. Then, a square mouth-centered RoI is extracted for each frame. To account for the yaw variations, the size of the RoI is set to the horizontal distance between the two mouth corners extended by an empirically determined factor of $0.12$, or twice the distance between the nose tip and the center of the mouth ($d_{MN}$), whichever is larger. However, this crop sometimes extends beyond the desired region for extremely small faces, so we restrict the size of the region to be no more than $3.2d_{MN}$.

In other words, the size of a RoI bounding box is determined by
\[w = \min \{3.2d_{MN}, \max \{2d_{MN}, 1.12x_r - 0.88x_l\} \},\]
where $x_l$ and $x_r$ are the $x$ coordinates of the left and right mouth corners. Finally, to smooth the resulting boxes, we apply a first-order Savitzky-Golay filter with window length $3$ to the estimated face rotations, the coordinates of the $x$ and $y$ centers, and the size of the ROIs.

\subsection{Validating the Extracted RoIs}
On some extremely challenging videos where the yaw and pitch angles are large, the landmark predictor fails and the extracted RoIs are inaccurate or even wrong. We train a binary CNN classifier to remove these non-lip images from the dataset. We begin the training process by using the initial unfiltered crops as positive samples and generate negative samples by shifting the crop region randomly in the original frame. After convergence, we filter the dataset using the trained model and fine-tune on the resulting subset. The trained model has a high recall (e.g. it easily picks up glasses at the top corner, and sometimes fails on profile views and low-resolution images, which are scarcer in the dataset), so we ask a human annotator to revise the inference results and remove false alarms.
  
\section{Dataset Statistics}
\emph{LRW-1000} is challenging due to its large variations in scale, resolution, background clutter, and speaker's attributes including pose, age, gender, make-up and so on. They are all important factors to consider when building a robust and practical lip-reading system. To facilitate the study of lip-reading, we provide cropped lip images and so users don't have to struggle with the trivial and cumbersome details of preprocessing. To quantify the properties of the datasets, we perform a comprehensive analysis based on the statistics of several aspects.

\subsection{Source Videos}
We select $51$ television programs with $840$ videos in total, where each raw video has a variable duration of $20$ minutes to $2$ hours. All the programs fall within the class of news and current affairs. Because different programs always have different broadcasters, we split all the videos of a single program into only one subset of the train, test, and validation set to ensure that there are no or few overlapped speakers among train, test and validation set. In summary, there are $840$ videos with about $508$ hours' duration in total. The principle of splitting train/test/validation follows two points:
(a) there are no or few overlapped speakers in these three sets; (b) the total duration of these three sets follows a ratio of about $8:1:1$, which means the number of samples in these three sets follows a similar ratio round $8:1:1$. 

Considering the above two points, we finally select $634$ videos of $44$ programs with more than $415$ hours for training, $84$ videos of $4$ programs with $43.4$ hours for test, and $122$ videos of $3$ programs with $48.95$ hours for validation.

\subsection{Word Samples}
In this subsection, we present the statistics about the word samples in this benchmark.

The final extracted samples in our benchmark have a duration of about $57$ hours with $718,018$ video clips in total, which are selected from the above $840$ raw videos. On the average, each class has about $718$ samples which makes it adequate to learn deep models. The minimum and maximum length of the data are  about $0.01$ seconds and $2.25$ seconds respectively, with an average of about $0.3$ seconds for each sample. This is reasonable as some words are indeed relatively short in our practical speaking process, especially the wake-up words in many speech-assistant systems. On the other hand, the abundance of such instances in real-world settings also suggests that they should not be overlooked in our research.

\subsection{Lip Region Resolution}
\begin{figure}[th]
  \centering
  \includegraphics[scale=0.5]{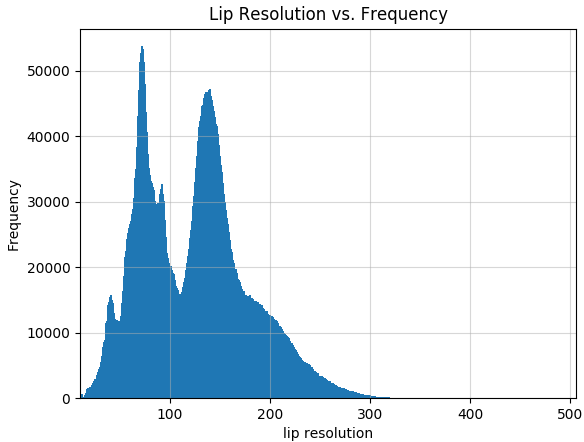}
  \setlength{\abovecaptionskip}{-0.1cm} 
  \setlength{\belowcaptionskip}{-0.5cm} 
  \caption{Scale distribution of the data, measured by the pixel-level width of the lip region.}  
  \vspace{-0.5cm}
  \label{fig_scale_distribution}
\end{figure}

Considering the diversity of scales of input videos in practical applications, we do not delete those sequences with small or large sizes. Instead, we collect these words according to their intrinsic natural distribution and found that there are indeed only few low or large sizes. Most of them contribute to some moderate value, as shown in Fig. \ref{fig_scale_distribution}. The two peaks in the figure are resulted by the two different types of the source videos: standard definition (SD) and high definition (HD). The existence of this case brings our benchmark much closer to practical applications which have a large resolution coverage.

\subsection{Speakers}
\begin{figure*} 
  \subfigure[Pitch Rotation Distribution]{ 
    \label{fig:mini:subfig:a} 
    \begin{minipage}[b]{0.3\textwidth} 
      \centering 
      \includegraphics[scale=0.3]{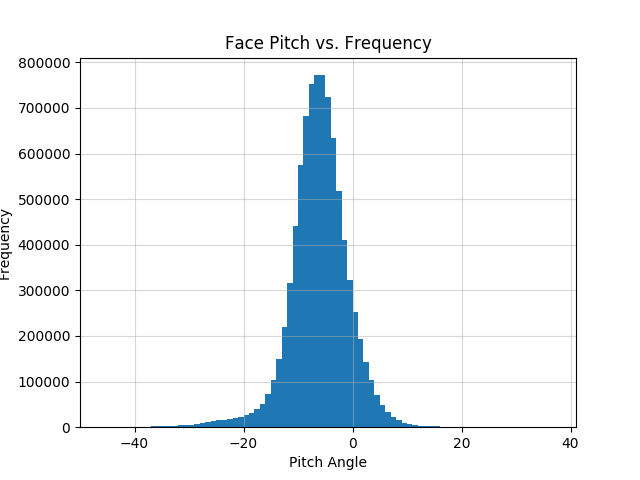} 
    \end{minipage}}%
  \subfigure[Yaw Rotation Distribution]{ 
    \label{fig:mini:subfig:b} 
    \begin{minipage}[b]{0.3\textwidth} 
      \centering 
      \includegraphics[scale=0.3]{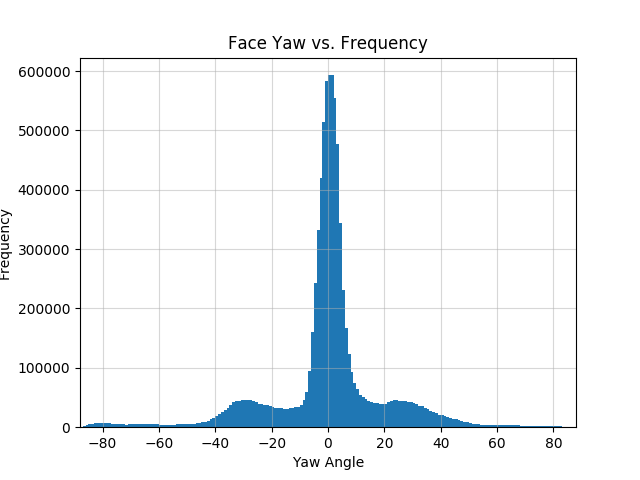} 
    \end{minipage}} 
  \subfigure[Roll Rotation Distribution]{ 
    \label{fig:mini:subfig:b} 
    \begin{minipage}[b]{0.3\textwidth} 
      \centering 
      \includegraphics[scale=0.3]{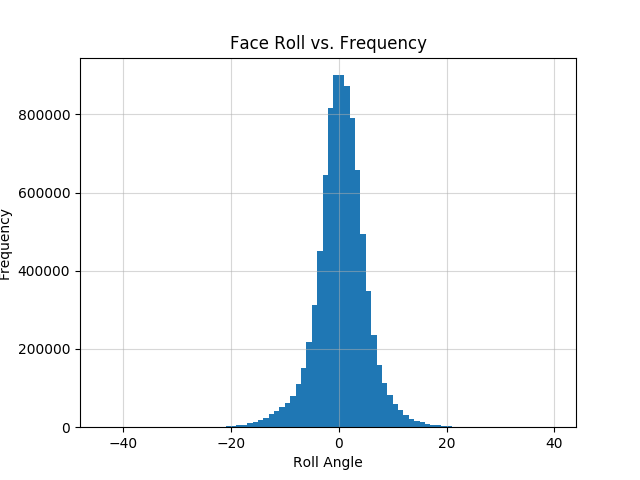} 
    \end{minipage}} 
  \caption{Pose distribution of the data in our benchmark, measured by angle degrees. Note that roll has been removed after preprocessing.} 
\label{fig_pose_distribution} 
\vspace{-0.6cm}
\end{figure*}

There are more than $2,000$ speakers in the $840$ videos used to construct our benchmark. The speakers are mostly interviewers, broadcasters, program guests, and so on. The large number and diversity in their identity equip the data with a broad coverage of age, pose, gender, accent, and personal speaking habits. These factors make the data very challenging for most existing lip-reading methods. We would evaluate the state-of-the-art word-level lip-reading models on our benchmark and the results should be very meaningful for designing practical lip-reading models. Among the multiple characteristics of speakers, we select pose as a statistical object because it is believed to be especially critical for the lip-reading task compared with other characteristics. We present the distribution of data in the pitch, yaw, and roll rotations respectively in Fig.~\ref{fig_pose_distribution}. We can see that although we do not perform deliberate filtering, the data is still mainly comprised of frontal views.

\section{Experiments}
In this section, we present the evaluation results of popular lip-reading methods and give a detailed analysis to illustrate the characteristics and challenges of the proposed benchmark.

\subsection{Baseline Methods}
We cast the word-level lip-reading task on our benchmark as a multi-class recognition problem and evaluate three popular methods on this dataset. Specifically, we evaluate three types of models with different types of front-end network: a fully 2D CNN based front-end, a fully 3D CNN based front-end and a front-end mixing 2D and 3D convolutional layers. Based on these three types of models, we hope to provide a relatively complete analysis and comparison of the currently popular methods. 

\begin{figure}[th]
  \centering
  \includegraphics[scale=0.9]{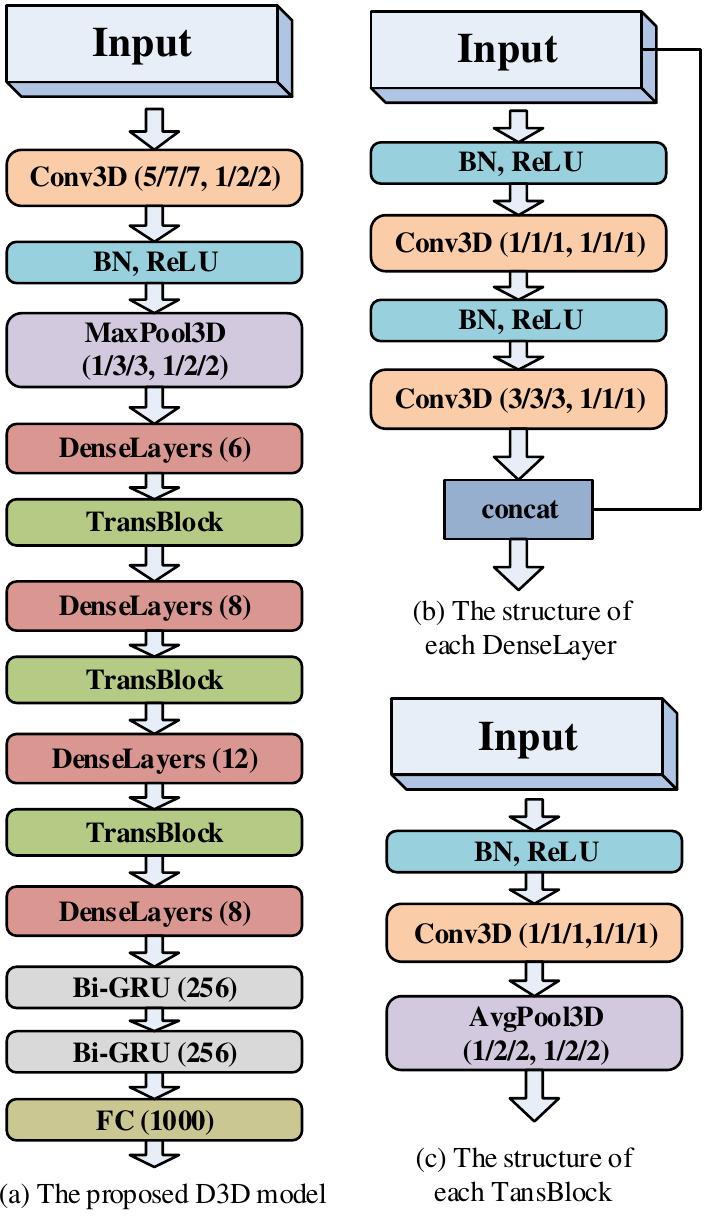}
  \setlength{\abovecaptionskip}{-0.1cm} 
  \setlength{\belowcaptionskip}{-0.5cm} 
  \caption{The proposed D3D network (DenseNet in 3D version).}
  \label{fig_d3d}
  \vspace{-0.8cm}
\end{figure}

The first network architecture in our experiments is the \emph{LSTM-5} network based on the multi-tower structure proposed in \cite{lstm5}, which is completely composed of 2D convolutional layers. This structure has achieved an appealing performance on the public word-level dataset \emph{LRW} \cite{VGG-LRW}. The second network is based on LipNet \cite{LIPNET} which contains only three spatio-temporal convolutional layers as the front-end. The third network is the model proposed in \cite{COMBINGING} which contains a 3D convolutional layer cascaded with a residual network as the front-end network. During the experiments, the original LipNet consistently failed to converge. We believe that this is because this dataset is too complex to be learned by only three spatio-temporal layers. Therefore, we propose to transform the 2D DenseNet into a 3D counterpart and apply it as the fully 3D convolutional front-end. We named this model as D3D (DenseNet in 3D version), whose structure is shown in Fig. \ref{fig_d3d}. These three models are abbreviated as ``LSTM-5'', ``3D+2D'' and ``D3D'' respectively in the experiments.

To perform a fair comparison, all the three models are combined with a back-end network of the same structure which contains a two-layer bi-directional RNN to perform the final recognition. The recurrent units used in our experiments are bidirectional Gated Recurrent Units. In the remainder of this section, we compare these three models side by side and also find some interesting observations.

\subsection{Experimental Settings}
1) \emph{Data Preprocessing: }
In our experiments, all the images are converted to grayscale and normalized with respect to the overall mean and variance. When fed into the models, the frames in each sequence are cropped in the same random position for training and centrally cropped for validation and test. All the images are resized to a fixed size of $122 \times 122$ and then cropped to a size of $112 \times 112$. As an effective data augmentation step, we also randomly flip all the frames in the same sequence horizontally.
To accelerate the training process, we divide the training process into two stages. In the first stage, we choose shorter sequences with a length below $30$, allowing a larger batch size for training. Then we add the remaining sequences to the training set when the models exhibit a tendency of convergence. We also randomly repeat some samples in the the training process to further accelerate the convergence.

2) \emph{Parameters Settings: }
Our implementation is based on PyTorch and the models are trained on servers with four NVIDIA Titan X GPUs, with 12GB memory of each one. We use the Adam optimizer with an initial learning rate of $0.001$, with $\beta=(0.9, 0.99)$. All the networks are pretrained on \emph{LRW}. During the training process, we apply dropout with probability $0.5$ to the last layer of each model to prevent the model from being trapped in some local optima for the \emph{LRW} dataset.

3) \emph{Evaluation Protocols: }
We provide two evaluation metrics in our experiments. The \textit{recognition accuracy} over all $1,000$ classes is naturally considered as the base metric, since this is a classification task. Meanwhile, motivated by the large diversity of the data shown in many aspects, such as the number of samples in each class, we also provide the \textit{Kappa coefficient} as a second evaluation metric.

\subsection{Recognition Results}

To evaluate the effects of different factors on lip-reading, we split the data into different difficulty levels according to the input scales (resolution), speaker's pose (degree), and the sample length (number of frames), as shown in Table~\ref{tab_split_easy_hard}. We now present a thorough comparison of the models on all three levels to obtain a complete and comprehensive analysis of the results. 

\begin{table}[th]
  \setlength{\abovecaptionskip}{-0.3cm} 
  \setlength{\belowcaptionskip}{-0.3cm} 
    \caption{Partition of Different Difficulty Levels on \textit{LRW-1000}}
    \label{tab_split_easy_hard}
    \begin{center}
        \begin{tabular}{|c||c|c|c|c|}
            \hline
            \textbf{Criterion} & \textbf{Easy} & \textbf{Medium} & \textbf{Hard}\\
            \hline
            \hline
            Input Scale  & $\leq$150 & $\leq$100 & $\leq$50\\            
            \hline            
            Pose  & $\geq$20 & $\geq$40 & $\geq$60\\    
            \hline
            Sample Length & $\leq$30 & $\leq$15 & $\leq$5\\   
            \hline
        \end{tabular}
    \end{center}
  \vspace{-0.7cm}
\end{table}

(1) \emph{General Performance:} 
We show the results on \emph{LRW} and \emph{LRW-1000} in Table \ref{tab_lrw_results} and Table \ref{tab_acc_results} respectively. We can see that there is a similar trend of these three models in both \emph{LRW} and \emph{LRW-1000}. The method combining 3D convolution together with 2D convolution performs best on both datasets. The \emph{LSTM-5} architecture, which relies only on the 2D convolutional layers performs worse compared to the other two models. This is reasonable because 3D convolution has an advantage for capturing short-term motion information, which has been proved important in lip-reading. However, the network with a fully 3D front-end cannot surpass the model combining 2D and 3D convolutional layers. This result proves the necessity of 2D convolutional layers for extracting fine-grained features in the spatial domain, which is quite useful for discriminating words with similar lip movements. In addition, the performance gap between these three models on \emph{LRW-1000} is not too wide and the Top-1 accuracy ranges from $25.76\%$ to $38.19\%$ among the $1,000$ classes, which confirms both the challenges and the consistency of our data.

\begin{table}[th]
  \setlength{\abovecaptionskip}{-0.3cm} 
  \setlength{\belowcaptionskip}{-0.2cm} 
    \caption{Recognition Results on \textit{LRW}}
    \label{tab_lrw_results}
    \begin{center}
        \begin{tabular}{|c||c|}
            \hline
            \textbf{Method} & \textbf{Accuracy}\\
            \hline
            \hline
            LSTM-5 & $66.0$\%\\
            \hline
            D3D & $78.0$\%\\
            \hline                
            3D+2D & $\mathbf{83.0\%}$\\
            \hline            
        \end{tabular}
    \end{center}
  \vspace{-0.5cm}
\end{table}

\begin{table}[th]
  \setlength{\abovecaptionskip}{-0.3cm} 
  \setlength{\belowcaptionskip}{-0.2cm} 
    \caption{Recognition Results on \textit{LRW-1000}}
    \label{tab_acc_results}
    \begin{center}
        \begin{tabular}{|c||c|c|c|c|}              
            \hline
            \textbf{Method} & \textbf{Top-1} & \textbf{Top-5} & \textbf{Top-10} & \textbf{Kappa (Top-1)}\\
            \hline
            \hline
            LSTM-5 & $25.76$\% & $48.74$\% & $59.73$\% & $0.24$ \\
            \hline
            D3D & $34.76$\% & $59.80$\% & $69.81$\% & $0.33$ \\
            \hline                
            3D+2D & $\mathbf{38.19\%}$ & $\mathbf{63.50\%}$ & $\mathbf{73.30\%}$ & $\mathbf{0.37}$ \\
            \hline            
        \end{tabular}
    \end{center}
  \vspace{-0.5cm}
\end{table}
(2) \emph{Performance vs. Word Length:} There is a small amount of samples with a relatively short duration in our benchmark, which can be used to roughly evaluate the performance of lip-reading models in extreme cases. The length is measured by the number of frames in our experiments. As shown in Fig.~\ref{fig_legnth_perf} and Table~\ref{tab_perf_length}, all models perform similarly when the word has a relatively short duration. As the length of the word gradually increases, the performance of all three models becomes better and more stable, likely because the context included in a sample increases simultaneously with the word's length. One other possible reason is that the number of samples within classes of longer durations is larger than those of shorter durations. 

\begin{figure}[th]
  \centering
  \includegraphics[scale=0.5]{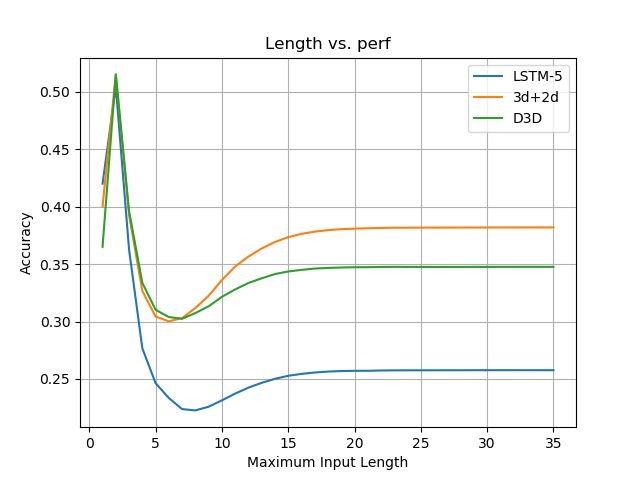}
  \setlength{\abovecaptionskip}{-0.3cm} 
  \setlength{\belowcaptionskip}{-0.5cm} 
  \setlength{\abovedisplayskip}{-13pt} 
  \caption{Recognition accuracy across word lengths (number of frames).}
  \label{fig_legnth_perf}
  \vspace{-0.3cm}
\end{figure}

\begin{table}[th]
  \setlength{\abovecaptionskip}{-0.3cm} 
  \setlength{\belowcaptionskip}{-0.3cm} 
    \caption{Performance w.r.t. Word Lengths (\# of frames) on \textit{LRW-1000}}
    \label{tab_perf_length}
    \begin{center}
        \begin{tabular}{|c||c|c|c||c|}
            \hline
            \textbf{Method} & \textbf{Easy} & \textbf{Medium} & \textbf{Hard} & \textbf{All}\\
            \hline
            \hline
            LSTM-5 & $25.76$\% & $25.27$\% & $24.63$\% & 25.76\%\\  
            \hline
            D3D & $34.75$\% & $34.36$\% & $\mathbf{31.01\%}$ & 34.76\% \\          
            \hline                
            3D+2D & $\mathbf{38.75\%}$ & $\mathbf{37.34\%}$ & $30.44$\% & $\mathbf{38.19\%}$\\
            \hline
        \end{tabular}
  \end{center}
  \vspace{-0.6cm}
\end{table}

\begin{figure}[th]
  \centering
  \includegraphics[scale=0.5]{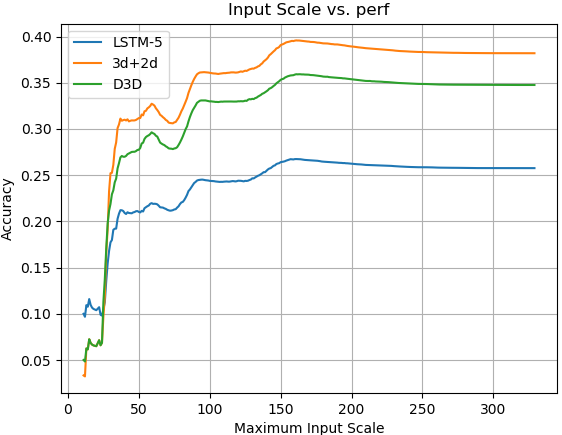}
  \setlength{\abovecaptionskip}{-0.1cm} 
  \setlength{\belowcaptionskip}{-0.5cm} 
  \caption{Recognition accuracy across input scales.}
  \label{fig_perf_resolution_total}
  \vspace{-0.5cm}
\end{figure}

(3) \emph{Performance vs. Input Scales: } 
We evaluate the models on the three levels which are divided by resolution, as shown in Table~\ref{tab_split_easy_hard}. Data with a resolution smaller than $50 \times 50$ falls in the hard level. Similarly, data with a resolution smaller than $100 \times 100$ and $150 \times 150$ fall in the medium level and the easy level, respectively. We can see that the performance of the models do tend to increase as we make a transition from the hard level to the medium level and from the medium level to the easy level. As shown in Table~\ref{tab_split_easy_hard} and Fig.~\ref{fig_perf_resolution_total}, the results show that higher input resolution does indeed help improve the lip-reading performance, but the performance would stabilize when the input scale is above some value. On the other hand, the performance gap between these three settings are not wide and the accuracy is still close to $30$\% for the $1,000$ classes even in the hard-level, where all the test sequences have a resolution below $50 \times 50$. This result again demonstrates the consistency of our data which covers a large variation over input scales.

\begin{table}[th]
  \setlength{\abovecaptionskip}{-0.3cm} 
  \setlength{\belowcaptionskip}{-0.5cm} 
    \caption{Performance w.r.t Input Scales on \textit{LRW-1000}}
    \label{tab_perf_scale}
    \begin{center}
        \begin{tabular}{|c||c|c|c||c|}
            \hline
            \textbf{Method} & \textbf{Easy} & \textbf{Medium} & \textbf{Hard} & \textbf{All} \\
            \hline
            \hline
            LSTM-5 & 26.41\% & 24.38\% & 21.02\% & 25.76\%\\   
            \hline
            D3D & 35.31\% & 32.98\% & 27.75\% & 34.76\%\\         
            \hline
            3D+2D & $\mathbf{39.08\%}$ & $\mathbf{36.07\%}$ & $\mathbf{31.18\%}$ & $\mathbf{38.19\%}$\\
            \hline
        \end{tabular}
    \end{center}
  \vspace{-0.3cm}
\end{table}

\begin{figure}[th]
  \centering
  \includegraphics[scale=0.5]{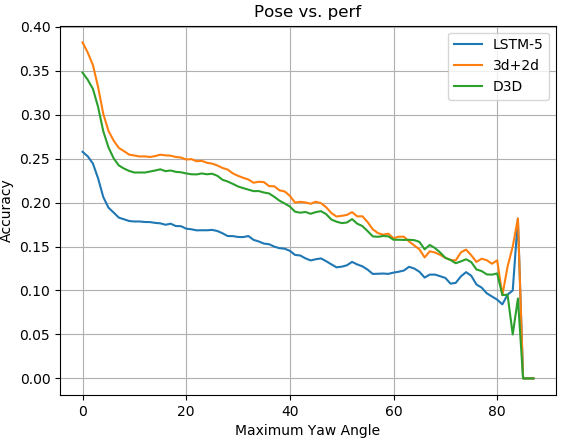}
  \setlength{\abovecaptionskip}{-0.1cm} 
  \setlength{\belowcaptionskip}{-0.5cm} 
  \caption{Recognition accuracy across poses (measured by yaw rotation).}
  \label{fig_perf_pose}
  \vspace{-0.5cm}
\end{figure}

(4) \emph{Performance vs. speaker pose: }
In this section, we evaluate the models under different poses measured by the yaw rotation. As shown in Fig.~\ref{fig_perf_pose} and Table~\ref{tab_perf_pose}, the performance of all three models drops greatly as the yaw angle increases. This may pose a serious challenge to most current lip-reading models in real-world scenarios. When speakers are viewed from a large angle, there is too much occlusion in the lip region, making it hard to learn the patterns from the data. This significant drop of performance when the camera viewpoints shift from frontal to profile may point out a challenging direction worthy of deeper study.

\begin{table}[th]
  \setlength{\abovecaptionskip}{-0.3cm} 
  \setlength{\belowcaptionskip}{-0.5cm} 
    \caption{Performance w.r.t Pose on \textit{LRW-1000}}
    \label{tab_perf_pose}
    \begin{center}
        \begin{tabular}{|c||c|c|c||c|}
            \hline
            \textbf{Method} & \textbf{Easy} & \textbf{Medium} & \textbf{Hard} & \textbf{All} \\
            \hline
            \hline
            LSTM-5 & 17.03\% & 14.51\% & 11.6\% & 25.76\%\\  
            \hline
            D3D & 23.31\% & 19.95\% & 15.78\% & 34.76\% \\          
            \hline            
            3D+2D & $\mathbf{24.89\%}$ & $\mathbf{20.76\%}$ & $\mathbf{15.9\%}$ & $\mathbf{38.19\%}$ \\
            \hline            
        \end{tabular}
    \end{center}
  \vspace{-0.6cm}
\end{table}

\addtolength{\textheight}{-3cm}   


\section{Conclusions}
In this paper, we have proposed a large-scale naturally-distributed word-level benchmark, named \emph{LRW-1000}, for lip-reading in the wild. We have evaluated representative lip-reading methods on our dataset to compare the effects of different factors on lip-reading. With this new dataset, we wish to present the community with some challenges of the lip-reading task -- scale, pose and word duration variations. These factors are ubiquitous in many real-world applications and very challenging for current lip-reading models. We look forward to new exciting research results inspired by the benchmark and the corresponding results provided in this paper.

\section{Acknowledgments}
This research was supported in part by the National Key R\&D Program of China (grant 2017YFA0700800), Natural Science Foundation of China (grants 61702486, 61876171). Shiguang Shan and Xilin Chen are corresponding co-authors of this paper.
\bibliography{ref_fg}
\end{document}